\documentclass[sigconf]{acmart}
\setcopyright{none}
\AtBeginDocument{%
  }

\setcopyright{acmlicensed}
\copyrightyear{2018}
\acmYear{2018}
\acmDOI{XXXXXXX.XXXXXXX}
\acmConference[Conference acronym 'XX]{Make sure to enter the correct
  conference title from your rights confirmation email}{June 03--05,
  2018}{Woodstock, NY}
\acmISBN{978-1-4503-XXXX-X/2018/06}

\usepackage{multirow}
\usepackage{pifont}

\usepackage{enumitem}  
\begin{document}

%%
%% The "title" command has an optional parameter,
%% allowing the author to define a "short title" to be used in page headers.
\title{ISPCloak: Weaponizing ISP for Optimization-Free Physical Camouflage against Deepfake Detectors}

%%
%% The "author" command and its associated commands are used to define
%% the authors and their affiliations.
%% Of note is the shared affiliation of the first two authors, and the
%% "authornote" and "authornotemark" commands
%% used to denote shared contribution to the research.
% \author{G.K.M. Tobin}
% \authornotemark[1]
% \email{webmaster@marysville-ohio.com}
% \affiliation{%
%   \institution{Institute for Clarity in Documentation}
%   \city{Dublin}
%   \state{Ohio}
%   \country{USA}
% }

\author{Jiale Zhao}
\affiliation{%
  \institution{Guangdong University of Technology}
  \city{Guangzhou}
  \country{China}
}
\email{zh2841871831@gmail.com}

\author{Jiajun Wan}
\affiliation{%
  \institution{Guangdong University of Technology}
  \city{Guangzhou}
  \country{China}
}
\email{892640097@mails.gdut.edu.cn}

\author{Lei Tang}
\affiliation{%
  \institution{Guangdong University of Technology}
  \city{Guangzhou}
  \country{China}
}
\email{3122002111@mail2.gdut.edu.cn}

\author{Ye Qin}
\affiliation{%
  \institution{Guangdong University of Technology}
  \city{Guangzhou}
  \country{China}
}
\email{3122000617@mail2.gdut.edu.cn}

\author{Kebing Jin}
\affiliation{%
  \institution{Guizhou Provincial Laboratory of Big Data, State Key Laboratory of Public Big Data, Guizhou University}
  %\institution{Guizhou Provincial Laboratory of Big Data}
  %\institution{State Key Laboratory of Public Big Data} 
  %\institution{Guizhou University}
  \city{Guiyang}
  \country{China}
}
\email{kbjin@gzu.edu.cn}

\author{Jinghui Qin}
\authornote{Corresponding author.}
\affiliation{%
  \institution{Guangdong University of Technology}
  \city{Guangzhou}
  \country{China}
}
\email{qinjinghui@gdut.edu.cn}
%%
%% By default, the full list of authors will be used in the page
%% headers. Often, this list is too long, and will overlap
%% other information printed in the page headers. This command allows
%% the author to define a more concise list
%% of authors' names for this purpose.

%%
%% The abstract is a short summary of the work to be presented in the
%% article.
\begin{abstract}
The rapid advancement of generative models has spurred the critical need to evaluate the worst-case robustness of deepfake detectors. In this paper, we reveal a fundamental blind spot in current forensic paradigms: while existing detectors excel at capturing digital synthesis artifacts, their effectiveness drops drastically when AI-generated content is cloaked in authentic physical imaging characteristics. We posit that genuine photographs inherently possess hardware-intrinsic statistical signatures, which are imperceptible footprints imprinted by optical sensors and Image Signal Processing (ISP) pipelines, and are fundamentally absent in purely data-driven generative models. Driven by this insight, we propose ISPCloak, a novel optimization-free adversarial attack framework that explicitly weaponizes the ISP pipeline to mislead the judgment of deepfake detectors. Rather than relying on computationally expensive gradient perturbations, our method first employs an Invertible ISP network to project images into the RAW domain. Then, we seamlessly imprint the complex statistical priors of real cameras onto AI-generated images by injecting realistic Poisson-Gaussian sensor noise and conducting forward ISP reconstruction. Synergized with generative artifact suppression and adaptive masking, this streamlined physical simulation enables ultra-fast generation of adversarial examples. Extensive experiments show that embedding authentic physical perturbations fundamentally disrupts a broad range of current detection mechanisms, yielding universally evasive adversarial examples with imperceptible visual alterations.
\end{abstract}

%%
%% The code below is generated by the tool at http://dl.acm.org/ccs.cfm.
%% Please copy and paste the code instead of the example below.
%%


%%
%% Keywords. The author(s) should pick words that accurately describe
%% the work being presented. Separate the keywords with commas.
%% A "teaser" image appears between the author and affiliation
%% information and the body of the document, and typically spans the
%% page.

% \received{20 February 2007}
% \received[revised]{12 March 2009}
% \received[accepted]{5 June 2009}

%%
%% This command processes the author and affiliation and title
%% information and builds the first part of the formatted document.
\maketitle

\newcommand{\wh}{\widehat}
\newcommand{\wt}{\widetilde}
\newcommand{\ov}{\overline}
\newcommand{\R}{\mathbb{R}}
%%% [A&C]: Paper comments command
\newcommand{\Jiale}[1]{{\color{red}[Jiale: #1]}} %%%Change to intern name

\section{Introduction}
\label{sec:intro}

\begin{figure*}[t!]
    \centering
    \includegraphics[width=0.95\textwidth]{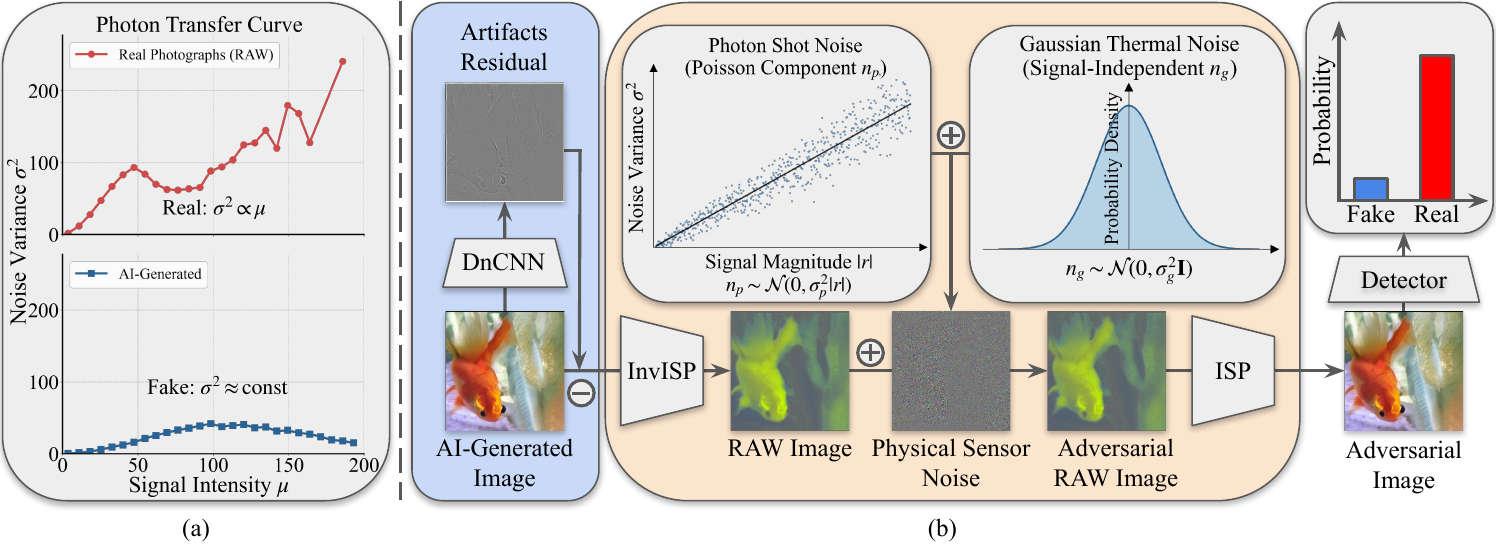}
    \caption{
        Overview of our proposed physical noise injection framework. (a) Motivation: Real photographs inherently follow Poisson photon transfer physics ($\sigma^2 \propto \mu$), whereas AI-generated images exhibit constant, physics-violating noise signatures. (b) Our Pipeline: Motivated by this gap, our framework suppresses synthetic artifacts via DnCNN and leverages an ISP cycle to inject authentic, signal-dependent photon and read noise in the RAW domain. This physically consistent injection yields highly stealthy and evasive adversarial examples that seamlessly emulate real sensor noise to bypass current detectors.
    }
    \label{fig:pipeline}
\end{figure*}

The rapid advancement of generative AI and deep learning has democratized the creation of photorealistic synthetic images. From text-to-image models like DALL-E~\cite{bgj+23} and Stable Diffusion~\cite{rbl+22} to face-swapping techniques and diffusion-based inpainting~\cite{zkr+25}, AI-generated content (AIGC) has proliferated at an unprecedented scale. This technological leap, while creatively empowering, admits a critical dual-use vulnerability. Specifically, AI-generated forgeries constitute a substantial threat to media authenticity, information integrity, and public trust. Thus, the forensic detection of AIGC has emerged as an urgent frontier, with numerous detection frameworks deployed to identify and mitigate synthetic imagery~\cite{gzu+25, wyf+25, llw+25, kta25, lll+25}. However, despite achieving impressive accuracy across various benchmarks, existing forensic methods fundamentally operate within a purely digital paradigm. Whether analyzing frequency spectra or extracting deep structural features, these frameworks primarily engage in an endless mathematical duel with generative models over the distributions of rendered pixels or latent representations. Consequently, by neglecting the hardware-intrinsic mechanics of natural image acquisition, these detectors remain fundamentally vulnerable to adversarial perturbations that accurately emulate authentic sensor signatures.

% The fundamental insight that motivates this work is deceptively simple yet profound. Real photographs are inextricably stamped with a unique, irreplaceable hardware-intrinsic statistical signature that no pure data-driven generative model can authentically replicate. When photons strike a CMOS or CCD sensor within a camera, they generate Poisson-distributed shot noise proportional to signal intensity. Simultaneously, thermal and readout electronics introduce Gaussian noise floors. These raw sensor readings then traverse a proprietary and highly nonlinear Image Signal Processing (ISP) pipeline, encompassing demosaicing, color interpolation, white balance, tone mapping, and denoising. This process imprints distinctive spatial correlations, chromatic cross-talk patterns, and frequency-domain characteristics unique to each camera model and sensor class. This complex transformation, rooted in physics and hardware design, produces an intricate statistical footprint across spatial, spectral, and inter-channel domains. AI-generative models, which learn from RGB distributions in pixel space or compressed latent spaces, fundamentally lack access to this underlying physical ground truth. They cannot authentically synthesize the RAW sensor domain or accurately emulate the nonlinear reshaping of sensor noise into naturalistic light reflections and textures by the ISP.

The fundamental insight that inspires this work is deceptively simple yet profound: \textit{real photographs are inextricably stamped with a unique, irreplaceable hardware-intrinsic statistical signature that no pure data-driven generative model can authentically replicate}. When photons strike a CMOS or CCD sensor within a camera, they generate Poisson-distributed shot noise proportional to signal intensity. Simultaneously, thermal and readout electronics introduce Gaussian noise floors. As empirically illustrated by the Photon Transfer Curve in Figure~\ref{fig:pipeline} (a), this fundamental electro-optical property dictates that the noise variance in real photographs inherently scales with signal intensity ($\sigma^2 \propto \mu$). These raw sensor readings then traverse a proprietary and highly nonlinear Image Signal Processing (ISP) pipeline, encompassing demosaicing, color interpolation, white balance, tone mapping, and denoising. This process imprints distinctive spatial correlations, chromatic cross-talk patterns, and frequency-domain characteristics unique to each camera model and sensor class. This complex transformation, rooted in physics and hardware design, produces an intricate statistical footprint across spatial, spectral, and inter-channel domains.

Conversely, AI-generative models, which learn from RGB distributions in pixel space or compressed latent spaces, fundamentally lack access to this underlying physical ground truth. As starkly contrasted in Figure~\ref{fig:pipeline} (a), these models fail to capture this dynamic physical relationship, defaulting instead to a physics-violating, constant noise variance profile ($\sigma^2 \approx \mathrm{const}$). They cannot authentically synthesize the RAW sensor domain or accurately emulate the nonlinear reshaping of sensor noise into naturalistic light reflections and textures by the ISP.

Existing adversarial defenses against AIGC detectors have largely overlooked this physical blind spot. Traditional gradient-based approaches, including Projected Gradient Descent (PGD)~\cite{mms+18}, alongside recent diffusion-based attack methods like Taigen~\cite{rjvs25} and Adv-diffusion~\cite{ldw+24}, operate exclusively in RGB or compressed latent spaces. These optimization-heavy methods suffer from several compounding limitations. First, their computational burden and latency render them impractical for large-scale deployment. Moreover, the perturbations manifest as artificial digital modifications or regimented adversarial noise, creating a categorical distinction from natural camera grain. Critically, because these digitally crafted perturbations rely heavily on model-specific gradients or latent representations, they are inherently overfitted to the feature space of the surrogate detectors. This severe overfitting fundamentally restricts their generalization capabilities, causing the adversarial effect to fail when transferring across heterogeneous forensic architectures or unseen detection paradigms in realistic black-box scenarios.

%To transcend this limiting paradigm, 
In this paper, we propose \textbf{ISPCloak}, a fundamentally novel and optimization-free physical adversarial attack framework that explicitly weaponizes the ISP pipeline to mislead the judgment of deepfake detectors. ISPCloak operates through a continuous, synergistic pipeline. First, we utilize a lightweight denoising network (e.g., DnCNN) to suppress inherent digital generative artifacts, yielding a purified image. Then, we employ a pre-trained invertible ISP network to project this purified image into the RAW sensor domain. Subsequently, within this domain, we introduce Poisson-Gaussian noise that explicitly models the physical photon and thermal characteristics of real camera hardware. To ensure strict imperceptibility, an adaptive gradient-based mask modulates these perturbations, deliberately allocating the physical noise into texture-rich regions while suppressing it in flat, smooth areas to preserve baseline visual fidelity. Finally, a forward ISP reconstruction organically transforms this RAW-domain noise into complex, naturalistic spatial and chromatic correlations. Remarkably, this entire pipeline is feedforward and parameter-free, involving no gradient descent or iterative optimization. An image can be processed in tens of milliseconds.

The physical grounding of ISPCloak produces three distinct advantages over existing methods. First, by operating in the RAW domain and leveraging true sensor physics, our perturbations are mathematically and perceptually indistinguishable from genuine camera noise. This endows the synthetic images with an authentic hardware fingerprint that forensic detectors cannot reliably distinguish from genuine captures. Second, the optimization-free design achieves near-real-time performance, enabling practical, scalable deployment in adversarial workflows. Third, the physical basis of our attack confers robust transferability. Because the RAW noise and ISP transformation are grounded in universal electromagnetic and optics principles, perturbations generated for one camera model remain effective across diverse sensor architectures. This intrinsic generalization property is fundamentally absent in purely digital attack paradigms.

The key contributions of this work are summarized as follows:
\begin{itemize}[noitemsep, topsep=0pt]
    \item We identify and formally characterize a critical physical blind spot in contemporary AIGC forensic detectors: the absence of authentic hardware-intrinsic statistical signatures in AI-generated images. This observation introduces a vital physical dimension to current forensic evaluations.
    
    \item We propose \textbf{ISPCloak}, the first optimization-free, physics-grounded adversarial attack framework that explicitly exploits the vulnerabilities of the ISP pipeline. By combining RAW-domain noise injection with adaptive masking and invertible ISP networks, we achieve imperceptible yet potent attacks that bypass traditional digital countermeasures.
    
    \item Extensive experiments validate the method's superiority across metrics of attack success rate (ASR), visual fidelity (PSNR/SSIM), computational efficiency, and cross-domain transferability. Our approach achieves significantly higher ASR than gradient-based and diffusion-based attacks while maintaining visual quality and operating orders of magnitude faster.
\end{itemize}

\section{Related Work}
\label{sec:related}

\subsection{Deepfake Detection}
Current Deepfake and AIGC detection frameworks primarily formulate the task as a binary classification of digital synthesis artifacts. While spatial methods target blending boundaries and textural anomalies~\cite{aiw25, llz+24}, frequency-domain approaches exploit abnormal distributions in Fourier or DCT spectra~\cite{tzw+24, ddn+25}. Furthermore, many data-driven detectors leverage CNNs or ViTs to identify architecture-specific noise or checkerboard patterns inherently tied to the generation process~\cite{thw+24}.
To enhance cross-domain generalization, recent literature has shifted toward more sophisticated feature extraction paradigms. One line of research focuses on structural and multi-domain inconsistencies: FatFormer~\cite{ltt+24} utilizes a forgery-aware adapter to fuse image and frequency features, whereas SAFE~\cite{lch+25} and NPR~\cite{tzs+24} specifically target local structural artifacts and neighboring pixel relationships introduced by up-sampling operations. Another emerging trend exploits vision-language priors; for instance, C2P-CLIP~\cite{ttl+25} injects category-related concepts into the image encoder to boost CLIP's detection potential, while AIDE~\cite{ylc+25} synergizes CLIP embeddings with multi-scale frequency patches. Additionally, approaches like LGrad~\cite{tzw+23} map images into a gradient space using pre-trained CNNs to extract domain-agnostic artifact representations.

Despite achieving remarkable performance in specific benchmarks, these defense mechanisms share a fundamental vulnerability: they remain heavily overfitted to the presence of digital synthetic traces, while entirely overlooking the absence of authentic physical traces. By operating within this purely digital paradigm, they fail to verify the hardware-intrinsic statistical signatures, including sensor noise patterns and ISP pipeline transformations, that universally govern authentic natural photographs. This paradigm-level blind spot leaves existing detectors highly susceptible to sophisticated attacks designed to convincingly emulate physical-world imaging characteristics.

\subsection{Adversarial Attacks on Deepfake Detectors}
To evaluate and expose the vulnerabilities of Deepfake detectors, various adversarial attack strategies have been proposed. Traditional adversarial attacks, such as the Fast Gradient Sign Method~\cite{gss14} and PGD~\cite{mms+18}, have been widely adapted to fool AIGC detectors by adding imperceptible, mathematically optimized noise perturbations to the image. To further enhance visual stealthiness and cross-model transferability, recent advancements have explored diffusion-based generation to bypass detection networks~\cite{ccc+24, hyw+25, hs25}. For instance, StealthDiffusion~\cite{zsc+24} optimizes the diffusion latent space to align the frequency spectra of adversarial and genuine images. Similarly, Diff-PGD~\cite{xahc23} incorporates diffusion-guided gradients into the traditional PGD framework, ensuring that perturbations strictly adhere to natural data distributions. Breaking away from conventional $L_p$-norm constraints, DiffAttack~\cite{ccc+24} injects semantic-level perturbations directly into the latent space, while DiffAdvMAP~\cite{pcz25} leverages diffusion priors to sample unrestricted adversarial examples from posterior distributions. Furthermore, to circumvent the inevitable distortions caused by post-processing existing images, the Adversarial Diffusion Model (ADM)~\cite{wlqz26} generates undetectable adversarial images entirely from scratch by optimizing an adversarial denoising U-Net and decoder to match the high-frequency characteristics of real data.

However, these dominant attack paradigms face severe practical limitations. First, their reliance on iterative backpropagation or sequential denoising renders them computationally prohibitive and unsuitable for real-time deployment. Second, the meticulously optimized perturbations heavily overfit to specific white-box surrogate models, causing a drastic drop in transferability against unseen black-box detectors. In contrast to these computationally heavy and mathematically overfitted digital attacks, our method introduces an optimization-free, single forward-pass pipeline that achieves high black-box transferability through universally applicable physical camouflage.

\subsection{Computational Photography and ISP}
The ISP pipeline is a fundamental component in digital cameras that bridges the gap between raw sensor measurements and the final perceivable RGB images~\cite{ucjk21, cm22}. When photons hit the Complementary Metal-Oxide-Semiconductor (CMOS) sensor, they induce a raw electrical signal containing signal-dependent Poisson noise (shot noise) and signal-independent Gaussian noise (read noise). The ISP transforms this raw, noisy signal through a complex, highly non-linear sequence of operations, including black level compensation, white balance, demosaicing, color space conversion, and gamma correction. This intricate process inherently weaves the raw sensor noise into complex, cross-channel, and spatially correlated textures in the final RGB domain, forming a distinct hardware-intrinsic statistical signature.

Historically, research in computational photography has focused on modeling these ISP pipelines and sensor noise distributions for positive visual objectives, such as image restoration, denoising, and enhancement~\cite{zzz+24,awzb25,wqy+25}. Recently, researchers have begun to integrate the ISP pipeline into adversarial attack workflows, recognizing its pivotal role in the physical-to-digital transition. To ensure the real-world robustness of physical perturbations, methods like ProjAttacker~\cite{lwj+25} and Camera-Agnostic Patch (CAP)~\cite{WWZ+24} incorporate differentiable ISP simulations into their optimization processes, effectively bridging the domain gap for projection- and patch-based attacks. Conversely, other works exploit the ISP's non-linear transformations to craft camera-specific attacks~\cite{pmh21}, demonstrating that specific hardware pipelines can selectively amplify adversarial patterns to deceive downstream classifiers. Furthermore, vulnerabilities within modern AI-driven ISPs have been exposed through data-poisoning backdoor attacks, such as the Neural Invisibility Cloak (NIC)~\cite{zjl+25}, which compromises the ISP to maliciously erase specific targets from the output images. Notably, to specifically bypass forensic media detectors, methods like SpoC~\cite{ctr+21} utilize a GAN-based approach to explicitly inject proprietary camera traces, such as those arising from demosaicing or compression, into synthetic images, deceiving detectors into believing the AI-generated media was acquired by a specific real-world camera model.

Unlike existing methods that use the ISP merely as a proxy for iterative optimization or rely on data-driven trace synthesis like SpoC~\cite{ctr+21} to superimpose empirical camera fingerprints entirely within the RGB domain, our work introduces an optimization-free, physical approach. Rather than simulating fingerprints at the surface level, we utilize an invertible ISP network to project AI-generated images back into the foundational RAW domain. By injecting authentic physical sensor noise and reconstructing the image through the forward ISP, we provide AIGC imagery with a universally evasive, physically grounded camouflage that prior empirical attacks cannot replicate.
\section{ISPCloak}
\label{sec:method}

\subsection{Overview}

The core paradigm shift of \textbf{ISPCloak} is rooted in a physical perspective. Rather than competing with detectors in an iterative optimization race that confines adversarial conflict to the mathematical manipulation of RGB pixels, we bypass learned decision boundaries by operating in the physical imaging domain. We posit that since generative models learn primarily from digital RGB distributions or compressed latents, they fundamentally lack the hardware-intrinsic signatures inherent to genuine camera captures. \textbf{ISPCloak} addresses this gap by projecting images into the RAW sensor domain to inject perturbations that follow the physical laws of photons and electronics. This creates a fundamental asymmetry: by imprinting authentic physical characteristics, we effectively mask the underlying generative traces and align the synthetic images with the natural photographic manifold, rendering them indistinguishable to detectors confined to the digital domain.

As illustrated in \figurename~\ref{fig:pipeline} (b), the complete \textbf{ISPCloak} pipeline is structured as a synergistic, optimization-free feedforward process. The workflow begins with generative artifact suppression, where high-frequency digital signatures inherent to AI-generated images are extracted and selectively attenuated. Subsequently, we employ a pre-trained invertible ISP network to perform an inverse mapping from the RGB space to the RAW sensor domain. Within this domain, we introduce Poisson-Gaussian noise that explicitly models the physical photon and thermal characteristics of actual camera hardware. Then, we employ an adaptive gradient-based mask to concentrate these physical perturbations in texture-rich regions while leveraging visual masking principles to maintain high visual fidelity. Finally, a forward ISP reconstruction transforms the physically grounded RAW features into complex spatial and chromatic correlations. %An overview of the complete framework is depicted in .

\subsection{Generative Artifact Suppression}

Prior research shows that AI-generated images carry distinctive digital artifacts as structural byproducts of synthesis. Examples include denoising trajectories in diffusion models and upsampling operations in GANs~\cite{ctc21,oyl23,wmw+24,zsc+24}. These anomalies constitute the primary feature space exploited by contemporary forensic detectors.

In light of this, we follow recent paradigms advocating for artifact suppression as a pre-processing step. We systematically attenuate existing generative markers before introducing authentic physical characteristics. This provides two primary benefits. First, it reduces the density of traceable fingerprints to lower initial detection risks. Second, it establishes a neutral baseline. By suppressing these underlying digital traces, the subsequent hardware noise injection operates on a clean foundation. This ensures that authentic physical signatures are not confounded by competing artifacts.

We leverage a residual learning strategy to isolate high-frequency artifacts in AI-generated images. Specifically, a pre-trained Denoising Convolutional Neural Network (DnCNN)~\cite{zzc+17} is used to predict the noise component directly. Given an input image $x$, the extracted artifact residual is obtained as:
% $
\begin{equation}
    R = \mathcal{F}(x),
\end{equation}
% $
where $\mathcal{F}(\cdot)$ denotes the DnCNN mapping that captures synthetic generation traces while preserving the underlying image content.

Directly modifying the entire image may degrade overall visual quality and drastically shift image statistics. To address this, we apply an adaptive gradient mask to modulate the suppression intensity. We first compute the gradient magnitude $G$ of the image:
\begin{equation}
G = \sqrt{(\nabla_h x)^2 + (\nabla_v x)^2 + \eta},
\end{equation}
where $\nabla_h$ and $\nabla_v$ denote the horizontal and vertical spatial gradient operators, and $\eta$ is a small constant added to ensure numerical stability. Then, we construct an adaptive mask $M$:
\begin{equation}
M = \left( \frac{G}{\max(G)} \right)^{\gamma}.
\end{equation}
where $\gamma$ controls the selectivity of the mask. Consequently, texture-rich regions and edges receive mask values near unity, while smooth regions receive values near zero. We apply this mask to the extracted residual through element-wise multiplication:
\begin{equation}
\tilde{R} = M \odot R.
\end{equation}

The final cleaned image $x'$ is obtained by subtracting the masked residual from the original image, followed by a clamping operation to ensure valid pixel ranges:
\begin{equation}
x' = \text{Clip}(x - \alpha \tilde{R}, 0, 1),
\end{equation}
where $\alpha \in [0, 1]$ is a blend weight that controls the intensity of the suppression. 

\begin{table*}[t]
\centering
\caption{ASR (\%) on the \textbf{GenImage} dataset. The best results are highlighted in bold, and the second-best results are underlined.}
\label{tab:genimage}
% 2. 压缩行高 (默认是 1.0，0.85 能显著压扁表格)
\renewcommand{\arraystretch}{0.85} 
% 3. 微微压缩列间距
\setlength{\tabcolsep}{4pt}
% \resizebox{\textwidth}{!}{
\begin{tabular}{l|l|cccccccc|c}
\toprule
Detector & Method & ADM & BigGAN & Glide & Midjourney & SD v4 & SD v5 & VQDM & Wukong & \textbf{Mean} \\
\midrule
\multirow{5}{*}{AIDE} 
 & PGD & 94.90 & \underline{91.40} & 77.50 & 80.10 & 67.50 & 71.00 & 86.40 & 76.60 & 80.67 \\
 & Diff-PGD & 90.50 & 58.60 & 57.30 & 68.80 & 44.10 & 42.30 & 70.30 & 54.90 & 60.85 \\
 & DiffAttack & 34.30 & 42.10 & 12.80 & 32.60 & 39.90 & 39.30 & 35.00 & 46.90 & 35.36 \\
 & StealthDiffusion & \underline{96.80} & \textbf{92.90} & \underline{82.20} & \underline{84.40} & \underline{79.10} & \underline{80.00} & \underline{89.40} & \textbf{87.20} & \underline{86.50} \\
 & Ours & \textbf{98.10} & 89.30 & \textbf{83.20} & \textbf{87.60} & \textbf{79.60} & \textbf{80.70} & \textbf{91.10} & \underline{83.10} & \textbf{86.59} \\
\midrule
\multirow{5}{*}{SAFE} 
 & PGD & 72.20 & \underline{89.20} & \underline{89.10} & 67.50 & 56.30 & 53.10 & 65.80 & 59.60 & 69.10 \\
 & Diff-PGD & \underline{77.20} & 82.10 & 79.60 & \underline{72.50}
 & \underline{65.30} & \underline{66.40} & \underline{74.60} & \underline{67.10} & \underline{73.10} \\
 & DiffAttack & 36.20 & 36.00 & 37.90 & 39.00 & 44.20 & 42.80 & 39.70 & 38.30 & 39.26 \\
 & StealthDiffusion & 21.00 & 18.70 & 21.10 & 22.50 & 25.40 & 23.70 & 25.10 & 22.40 & 22.49 \\
 & Ours & \textbf{84.90} & \textbf{96.20} & \textbf{95.90} & \textbf{80.20} & \textbf{70.60} & \textbf{68.20} & \textbf{83.00} & \textbf{73.30} & \textbf{81.54} \\
\midrule
\multirow{5}{*}{C2P-CLIP} 
 & PGD & \underline{76.60} & 26.90 & \textbf{90.40} & \underline{98.60} & \underline{89.70} & \underline{89.20} & \underline{62.20} & \underline{87.70} & \underline{77.66} \\
 & Diff-PGD & \textbf{80.50} & \textbf{32.60} & 78.60 & 93.60 & 82.00 & 80.40 & 57.30 & 80.80 & 73.22 \\
 & DiffAttack & 12.60 & 19.20 & 16.50 & 35.00 & 42.80 & 44.30 & 47.00 & 27.30 & 30.59 \\
 & StealthDiffusion & 56.10 & 16.50 & 65.60 & 93.00 & 81.30 & 82.50 & 48.60 & 72.10 & 64.46 \\
 & Ours & 75.10 & \underline{29.60} & \underline{87.20} & \textbf{98.60} & \textbf{96.60} & \textbf{94.60} & \textbf{73.20} & \textbf{94.70} & \textbf{81.20} \\
\midrule
\multirow{5}{*}{LGrad} 
 & PGD & 41.00 & 47.70 & 33.20 & 30.00 & 44.00 & 42.60 & 37.30 & 49.80 & 40.70 \\
 & Diff-PGD & 30.20 & 26.30 & 21.20 & 18.40 & 28.80 & 28.50 & 28.10 & 30.30 & 26.47 \\
 & DiffAttack & 47.40 & \textbf{65.30} & 39.00 & 35.20 & \underline{52.00} & \underline{50.40} & \underline{41.50} & \underline{51.70} & 47.81 \\
 & StealthDiffusion & \underline{50.60} & 56.10 & \textbf{50.00} & \underline{46.20} & 48.90 & 49.10 & 40.00 & 50.60 & \underline{48.94} \\
 & Ours & \textbf{59.50} & \underline{64.10} & \underline{45.60} & \textbf{49.00} & \textbf{65.40} & \textbf{66.90} & \textbf{53.40} & \textbf{68.90} & \textbf{59.10} \\
\bottomrule
\end{tabular}
% }
\end{table*}

%\subsection{Invertible ISP Mapping and Physical Noise Synthesis}
\subsection{Physical Noise Injection and Invertible ISP}
The critical distinction between our approach and traditional adversarial methods lies in operating within the RAW photonic domain rather than the processed RGB color space. To achieve this, we leverage an Invertible Image Signal Processing (InvISP) framework~\cite{xqc21}, which enables highly accurate bidirectional mapping between RAW and RGB domains via learned neural transformations.

Given the cleaned RGB image $x'$ from the previous stage, we perform inverse ISP mapping to obtain its RAW representation:
\begin{equation}
r = \Phi^{-1}(x'),
\end{equation}
where $\Phi^{-1}$ denotes the learned inverse ISP transformation parameterized by the InvISP network. This function maps the 3-channel RGB data back to a Bayer-mosaiced RAW format. Then, the Bayer-mosaiced RAW data will be injected with physical sensor noise that models fundamental photon transport and thermal electronics principles. The injected noise comprises two additive components:
\begin{equation}
r' = r + n_p + n_g.
\end{equation}

The Poisson component $n_p$ models photon shot noise, which is inherent to the stochastic arrival of photons at the photodiode~\cite{ftke08, bmx+19}. Following fundamental sensor physics, the variance of shot noise is proportional to the signal magnitude. Simultaneously, we consider Gaussian thermal noise $n_g$ arising from thermal energy in the readout electronics and sensor substrate. For computational tractability, we approximate the Poisson shot noise using a signal-dependent Gaussian distribution, yielding the following unified noise formulations:
\begin{equation}
\begin{aligned}
    n_p & \sim \mathcal{N}(0, \sigma_p^2 |r|), \\ 
    n_g & \sim \mathcal{N}(0, \sigma_g^2 \mathbf{I}),
\end{aligned}
\end{equation}
where $\sigma_p$ and $\sigma_g$ are tunable scaling coefficients governing the magnitudes of the signal-dependent shot noise and the signal-independent thermal noise, respectively. The variance dependency on signal intensity for $n_p$ represents the fundamental signature of Poisson statistics, where brighter pixels naturally exhibit higher variance due to increased photon flux.

To prevent excessive signal clipping or unrealistic noise magnitudes, we constrain the perturbation via a clipping function:
\begin{equation}
\hat{r} = \text{Clip}(r', r - \epsilon, r + \epsilon),
\end{equation}
where $\epsilon$ bounds the RAW value perturbation to physically realistic sensor ranges.

The final step of this core stage is image reconstruction with forward ISP, wherein the noisy RAW image undergoes the learned forward transformation to produce the final adversarial image:
% $
\begin{equation}
    x^* = \Phi(\hat{r}),
\end{equation}
% $
where $\Phi$ represents the forward ISP mapping. This transformation is crucial because the ISP pipeline encompasses multiple non-linear processing stages, including demosaicing, color interpolation, white balance adjustment, and gamma correction. When applied to the injected RAW noise, these sensor-specific operations introduce complex spatial correlations and inter-channel interference. The noise that was independent in the RAW space becomes intricately woven into the RGB image globally. Generative models operating in pixel or latent spaces cannot reproduce this transformation because they lack access to the physical ground truth. Consequently, the resulting adversarial image $x^*$ carries an authentic hardware fingerprint. Forensic detectors trained on real camera imagery struggle to distinguish this statistical signature from genuine captures.

\begin{table*}[t]
\centering
\caption{ASR (\%) on \textbf{WildFake} dataset. The best results are highlighted in bold, and the second-best results are underlined.}
\label{tab:wildfake}
% 2. 压缩行高 (默认是 1.0，0.85 能显著压扁表格)
\renewcommand{\arraystretch}{0.85} 
% 3. 微微压缩列间距
\setlength{\tabcolsep}{4pt}
% \resizebox{\textwidth}{!}{
\begin{tabular}{l|l|cccccccc|c}
\toprule
Detector & Method & ADM & DALL-E & DDIM & DDPM & Imagen & Midjourney & SD & VQDM & \textbf{Mean} \\
\midrule
\multirow{5}{*}{AIDE} 
 & PGD & 94.30 & 67.10 & 92.30 & 92.80 & 37.40 & 73.30 & 67.00 & \underline{84.80} & 76.12 \\
 & Diff-PGD & 90.50 & 54.10 & 87.50 & 90.60 & 22.90 & 55.50 & 50.60 & 74.60 & 65.79 \\
 & DiffAttack & 33.50 & 44.70 & 31.00 & 29.50 & 35.50 & 45.20 & 49.80 & 40.10 & 38.66 \\
 & StealthDiffusion & \underline{96.60} & \underline{90.70} & \underline{97.10} & \underline{97.00} & \underline{86.60} & \underline{93.60} & \textbf{93.20} & 64.00 & \underline{89.85} \\
 & ISPCloak (Ours) & \textbf{99.90} & \textbf{98.60} & \textbf{100.00} & \textbf{99.90} & \textbf{96.50} & \textbf{93.60} & \underline{88.00} & \textbf{99.90} & \textbf{97.05} \\
\midrule
\multirow{5}{*}{SAFE} 
 & PGD & \underline{76.00} & \underline{74.30} & \underline{79.10} & \underline{67.80} & \underline{77.40} & \underline{72.50} & \underline{69.00} & \underline{70.80} & \underline{73.36} \\
 & Diff-PGD & 65.80 & 70.90 & 70.60 & 63.60 & 70.80 & 62.80 & 61.70 & 49.70 & 64.49 \\
 & DiffAttack & 39.50 & 42.20 & 38.50 & 42.80 & 42.10 & 36.20 & 39.40 & 37.00 & 39.71 \\
 & StealthDiffusion & 20.80 & 17.40 & 22.20 & 23.60 & 20.20 & 17.90 & 20.70 & 23.10 & 20.74 \\
 & ISPCloak (Ours) & \textbf{100.00} & \textbf{98.90} & \textbf{100.00} & \textbf{99.70} & \textbf{100.00} & \textbf{99.80} & \textbf{99.70} & \textbf{99.80} & \textbf{99.74} \\
\midrule
\multirow{5}{*}{C2P-CLIP} 
 & PGD & 74.70 & \underline{83.70} & \underline{48.30} & \underline{55.20} & \underline{96.30} & \underline{75.80} & \underline{88.70} & \underline{62.20} & \underline{73.11} \\
 & Diff-PGD & \underline{79.20} & 83.10 & 36.00 & 38.20 & 87.80 & \textbf{78.40} & 87.20 & 49.20 & 67.39 \\
 & DiffAttack & 12.80 & 47.10 & 38.30 & 42.80 & 31.10 & 11.70 & 23.90 & 47.20 & 31.86 \\
 & StealthDiffusion & 61.70 & 83.10 & 37.50 & 38.50 & 91.30 & 73.90 & \textbf{89.70} & 53.50 & 66.15 \\
 & ISPCloak (Ours) & \textbf{89.70} & \textbf{85.60} & \textbf{58.90} & \textbf{55.30} & \textbf{96.50} & 62.20 & 84.30 & \textbf{68.20} & \textbf{75.09} \\
\midrule
\multirow{5}{*}{LGrad} 
 & PGD & 40.60 & \underline{62.10} & 43.40 & 31.80 & 24.90 & 39.60 & 63.30 & 40.50 & 43.27 \\
 & Diff-PGD & 25.80 & 46.30 & 52.40 & 39.30 & 19.60 & 17.10 & 46.50 & 23.00 & 33.75 \\
 & DiffAttack & 44.10 & 59.30 & 47.40 & 44.40 & 28.80 & \underline{43.80} & 61.40 & 40.50 & 46.21 \\
 & StealthDiffusion & \underline{45.60} & \textbf{62.70} & \underline{55.80} & \underline{44.90} & \textbf{46.20} & \textbf{49.00} & \textbf{66.00} & \underline{41.00} & \underline{51.40} \\
 & ISPCloak (Ours) & \textbf{63.00} & 59.10 & \textbf{73.30} & \textbf{57.60} & \underline{33.20} & 30.20 & \underline{65.40} & \textbf{57.10} & \textbf{54.86} \\
\bottomrule
\end{tabular}
% }
\end{table*}

\section{Experiments}\label{sec:exp}

\subsection{Experimental Settings}

\subsubsection{Datasets.} We evaluate our framework on three complementary datasets chosen to highlight distinct challenges in modern forgery detection. GenImage~\cite{zcy+23} contains large-scale full-image synthesis from diffusion and GAN models, providing a challenging setting where global digital artifacts appear without sensor-consistent acquisition. WildFake~\cite{hz24} comprises unconstrained real-world images from diverse sources, testing the transferability of our physically grounded perturbations across heterogeneous distributions. FaceForensics++~\cite{rcv+19} focuses on localized facial manipulations, allowing us to evaluate whether ISP-based noise can homogenize statistics between authentic backgrounds and manipulated regions.

\subsubsection{Detectors.} We benchmark ISPCloak against 4 diverse forensic detectors under a transfer-based black-box setting. C2P-CLIP~\cite{ttl+25} and AIDE~\cite{ylc+25} leverage Vision-Language Models with cross-modal cues and frequency priors. SAFE~\cite{lch+25} targets spatial and structural inconsistencies, particularly from local artifacts or up-sampling. LGrad~\cite{tzw+23} captures domain-agnostic gradient representations. This selection covers complementary forensic cues, enabling evaluation of whether our physical perturbations consistently challenge structural, frequency, gradient, and cross-modal signals.

\subsubsection{Baselines.} ISPCloak is compared with four representative adversarial strategies. PGD~\cite{mms+18} is a classical iterative $L_p$-norm constrained attack. Diff-PGD~\cite{xahc23} and StealthDiffusion~\cite{zsc+24} combine gradient-based optimization with generative guidance. DiffAttack~\cite{ccc+24} performs latent-space generative perturbations aligned with frequency-domain characteristics. Unlike these optimization-heavy baselines that risk overfitting surrogate models, ISPCloak uses a single, forward-pass pipeline, achieving robust black-box transferability with minimal computation.

% \subsubsection{Evaluation Metrics.}
% We evaluate our framework from two complementary perspectives: evasion effectiveness and visual fidelity. ASR quantifies the fraction of adversarial images that successfully bypass the target detector under a transfer-based black-box setting. To rigorously assess visual fidelity and perturbation imperceptibility, we report Peak Signal-to-Noise Ratio (PSNR) for pixel-level fidelity, Structural Similarity Index (SSIM) for structural integrity, and $L_2$ distance for overall pixel deviation. Additionally, we employ Fréchet Inception Distance (FID) to quantify the statistical similarity between the deep feature distributions of real and adversarial images, ensuring that the generated perturbations remain natural and semantically consistent.

\subsubsection{Evaluation Metrics.} Evasion effectiveness is quantified by the ASR metric in black-box scenarios. To rigorously assess visual and statistical fidelity, we report PSNR, SSIM, $L_2$ distance, and FID, ensuring that the generated perturbations remain imperceptible and semantically natural.

\subsubsection{Implementation Details.} All experiments are conducted on an NVIDIA RTX 4090D. To ensure a balanced and large-scale evaluation, we randomly sample $1,000$ images from each generative sub-category for GenImage and WildFake, resulting in $8,000$ images respectively. For FF++, we randomly select images across all 4 manipulation methods. This strategy is designed to maintain the independence of video sources, ensuring that each sample originates from a distinct underlying sequence to maximize the diversity of identities and backgrounds, while avoiding the content redundancy inherent in per-category sampling. All images are maintained at their original resolutions with a fixed seed of $42$. For our ISPCloak, the noise scaling coefficients are set to $\sigma_p = 0.09$ and $\sigma_g = 0.075$. The perturbations are strictly clipped at $\epsilon = 0.006$, with blending factor $\alpha = 0.2$ and mask power $\gamma = 2.5$. 
To facilitate community verification and comparative evaluation, the original images along with their corresponding adversarial examples will be released upon publication.

To evaluate the black-box transferability of baselines, we intentionally select a wide range of heterogeneous surrogate models, including CNNs and Transformers. This diverse selection allows us to rigorously assess whether adversarial perturbations can generalize across fundamentally different architectural priors. Specifically, for PGD, we use a ResNet-50~\cite{hzrs16} surrogate with $\epsilon = 0.03$ and a step size of $0.0039$ over $30$ iterations. For Diff-PGD, a Swin-Transformer~\cite{llc+21} is employed, utilizing $6$ iterations with step size $1.0$ and $3$ diffusion steps under a DDIM-50 schedule. DiffAttack utilizes the Data-efficient Image Transformer (DeiT)~\cite{tcd21} as its surrogate, performing $10$ optimization iterations with $10$ diffusion steps. Finally, for StealthDiffusion, we employ an EfficientNet~\cite{tl19} surrogate and follow the settings in~\cite{zsc+24}, with $5$ iterations and $2$ latent diffusion steps at $\epsilon = 0.0157$.

\begin{table}[ht]
\centering
\caption{ASR (\%) on \textbf{FaceForensics++} dataset. The best results are highlighted in \textbf{bold}.}
\label{tab:ff}
% 2. 压缩行高 (默认是 1.0，0.85 能显著压扁表格)
\renewcommand{\arraystretch}{0.85} 
% 3. 微微压缩列间距
\setlength{\tabcolsep}{4pt}
\begin{tabular}{lcccc}
\toprule
Method & AIDE & SAFE & C2P-CLIP & LGrad \\
\midrule
PGD & 45.20 & 90.80 & 58.70 & 60.50 \\
Diff-PGD & 43.20 & 84.80 & 31.50 & 56.20 \\
DiffAttack & 17.00 & 23.70 & 23.50 & 77.20 \\
StealthDiffusion & 58.70 & 15.70 & 37.60 & 75.20 \\
ISPCloak (Ours) & \textbf{82.60} & \textbf{95.60} & \textbf{64.50} & \textbf{87.70} \\
\bottomrule
\end{tabular}
\end{table}

\subsection{Main Results and Analysis}
\label{sec:main_results}

We evaluate the ASR of our method across three diverse datasets and four heterogeneous detectors. Overall, our approach proves highly effective across different generative sources and detection architectures.

As shown in Table~\ref{tab:genimage}, our method attains the highest mean ASR across all evaluated detectors on GenImage. Specifically, it achieves 86.59\% on AIDE, slightly surpassing the 86.50\% attained by StealthDiffusion. The advantage over prior methods is particularly evident on SAFE and C2P-CLIP, where our method achieves substantially higher success rates. In addition, our approach remains highly effective across nearly all generators, including challenging diffusion-based models such as SD v4, SD v5, and VQDM, indicating robustness to variations in generative pipelines.

A key observation from GenImage is the stability of our method across detectors with fundamentally different mechanisms. Generative attacks such as StealthDiffusion perform competitively on AIDE by reaching 86.50\%, but their efficacy drops to 22.49\% on SAFE and becomes less consistent on C2P-CLIP with a mean ASR of 64.46\%. Similarly, optimization-based methods such as PGD and Diff-PGD exhibit notable variance across generators and detectors. While prior works report higher success rates under specific surrogate-target pairs, our evaluation shows that under more diverse architectural gaps, the effectiveness of generative attacks degrades substantially. This behavior suggests sensitivity to the surrogate model and limited cross-detector generalization.

In contrast, our method demonstrates more stable transferability across these heterogeneous settings. On LGrad, our method achieves the highest mean ASR of 59.10\%, outperforming StealthDiffusion at 48.94\% and DiffAttack at 47.81\%. Similar trends are observed across most generators, where our approach consistently ranks among the top-performing methods. These results suggest that avoiding explicit gradient-based optimization and instead leveraging ISP-consistent perturbations improves robustness under cross-architecture evaluation.

Table~\ref{tab:wildfake} demonstrates the results on WildFake. Here, our method demonstrates top-performing capabilities and achieves the highest mean ASR on all evaluated defenses. In particular, the ASR reaches 97.05\% on AIDE and 99.74\% on SAFE, while improvements on the remaining detectors are consistent with those observed on GenImage. These results suggest that the proposed perturbation strategy generalizes exceptionally well to unconstrained, in-the-wild distributions.

Finally, Table~\ref{tab:ff} shows the performance on FaceForensics++. Prior approaches exhibit noticeable degradation in this localized manipulation setting. For instance, DiffAttack achieves only 23.50\% ASR on C2P-CLIP, whereas our method reaches 64.50\%. These findings establish that our ISPCloak decisively outperforms existing attacks. Specifically, our physically grounded perturbations demonstrate superior robustness in spatially heterogeneous spoofing contexts, seamlessly masking the structural boundaries between manipulated regions and authentic backgrounds.

Ultimately, the comprehensive evaluation across all 3 datasets confirms that our ISPCloak provides highly effective and consistent improvements over existing attacks, particularly under heterogeneous detector architectures and realistic testing conditions. This proves that physically motivated perturbations offer a fundamentally more robust and transferable alternative to purely optimization-driven attack strategies.

\begin{figure}[ht]
  \centering
  \includegraphics[width=\linewidth]{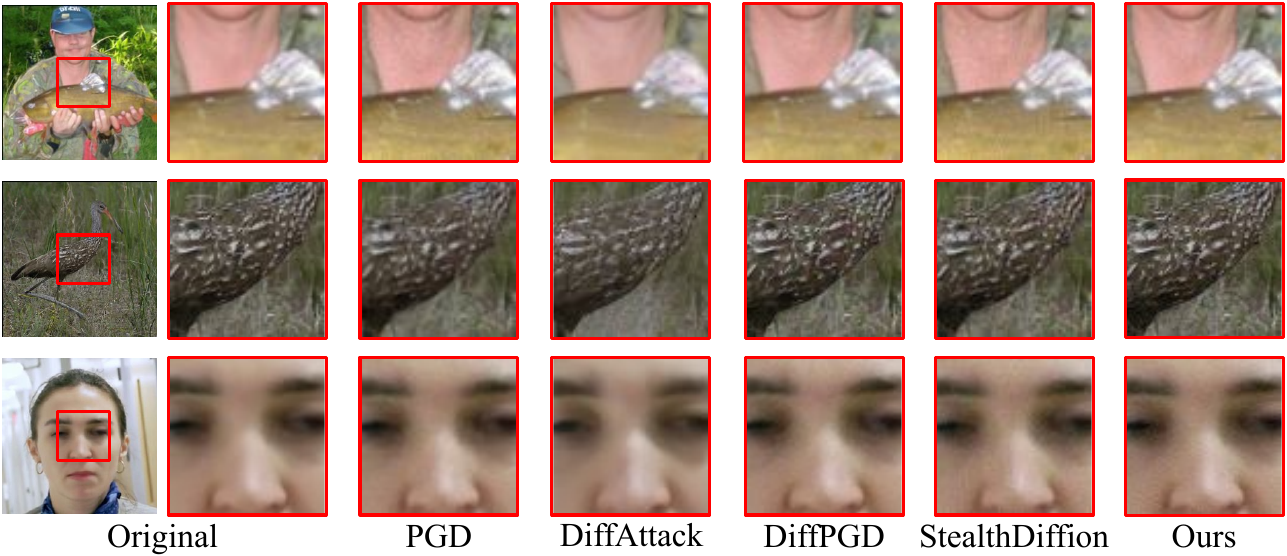}
    \caption{
    Qualitative comparison of adversarial examples across GenImage, WildFake, and FaceForensics++. Red boxes denote regions enlarged by $3.33\times$ to reveal details.
    }
  \label{fig:adversarial_comparison}
\end{figure}

\begin{table}[t]
\centering
\caption{Visual fidelity and image quality assessment on the \textbf{GenImage} dataset. The best results are highlighted in \textbf{bold}.}
\label{tab:image_quality}
\renewcommand{\arraystretch}{0.85} 
% 3. 微微压缩列间距
\setlength{\tabcolsep}{4pt}
\resizebox{\columnwidth}{!}{
\begin{tabular}{l cccc}
\toprule
\textbf{Method} & \textbf{PSNR} $\uparrow$ & \textbf{SSIM} $\uparrow$ & \textbf{$L_2$ Dist.} ($\times 10^{-2}$) $\downarrow$ & \textbf{FID} $\downarrow$ \\
\midrule
PGD & 33.82 & 0.86 & 2.53 & 19.17 \\
DiffAttack & 31.45 & 0.87 & 2.49 & 18.80 \\
Diff-PGD & 32.67 & 0.90 & 2.33 & 19.36 \\
StealthDiffusion & 33.58 & 0.88 & 2.12 & 20.95 \\
ISPCloak (Ours) & \textbf{35.56} & \textbf{0.91} & \textbf{1.68} & \textbf{14.21} \\
\bottomrule
\end{tabular}
}
\end{table}

\subsection{Visual Fidelity and Image Quality}

As shown in Fig.~\ref{fig:adversarial_comparison}, the perturbations introduced by ISPCloak exhibit visually natural characteristics that resemble sensor-level noise in real imaging systems. These perturbations follow the physical properties of photon shot noise and electronic read noise, resulting in images that remain perceptually consistent with the original content while effectively concealing generative artifacts.

The visual fidelity and image quality metrics on the GenImage dataset are reported in Table~\ref{tab:image_quality}. Our ISPCloak consistently achieves the best performance across all metrics, including PSNR, SSIM, $L_2$ distance, and FID. These results indicate that ISPCloak preserves high visual quality while effectively implementing adversarial perturbations, outperforming baseline attacks such as PGD, DiffAttack, Diff-PGD, and StealthDiffusion.

\begin{table}[ht]
\centering
\caption{Component ablation study on the GenImage dataset. We report the ASR evaluated against the SAFE detector alongside visual quality metrics.}
\label{tab:ablation}
\resizebox{\columnwidth}{!}{
\begin{tabular}{c|c|c|c|c|c ccccc}
\toprule
\multicolumn{6}{c}{\textbf{Components}} & \multicolumn{5}{c}{\textbf{Metrics}} \\
\cmidrule(lr){1-6} \cmidrule(lr){7-11}
$n_p$ & $n_g$ & DnCNN & Clip & Mask & ISP & \textbf{ASR} (\%)  & \textbf{PSNR}  & \textbf{SSIM}  & \textbf{$L_2$} ($\times 10^{-2}$)  & \textbf{FID}  \\
\midrule
           &            & \checkmark & \checkmark & \checkmark &            & 38.37 & 53.04 & 1.00 & 0.23 & 0.21 \\
\checkmark & \checkmark & \checkmark & \checkmark & \checkmark &            & 62.73 & 44.32 & 0.99 & 0.61 & 1.67 \\
           &            & \checkmark & \checkmark & \checkmark & \checkmark & 39.27 & 51.53 & 1.00 & 0.27 & 0.18 \\
           & \checkmark & \checkmark & \checkmark & \checkmark & \checkmark & 80.61 & 35.50 & 0.91 & 1.70 & 9.47 \\
\checkmark &            & \checkmark & \checkmark & \checkmark & \checkmark & 46.29 & 35.72 & 0.93 & 1.64 & 9.87 \\
\checkmark & \checkmark &            & \checkmark & \checkmark & \checkmark & 77.41 & 39.71 & 0.96 & 1.06 & 4.91 \\
\checkmark & \checkmark & \checkmark &            & \checkmark & \checkmark & 99.26 & 13.93 & 0.22 & 20.31 & 121.43 \\
\checkmark & \checkmark & \checkmark & \checkmark &            & \checkmark & 74.04 & 34.07 & 0.90 & 2.01 & 8.35 \\
\midrule
\checkmark & \checkmark & \checkmark & \checkmark & \checkmark & \checkmark & 81.54 & 35.56 & 0.91 & 1.68 & 14.21 \\
\bottomrule
\end{tabular}
}
\end{table}

\begin{table}[t]
\centering
\caption{Hyperparameter sensitivity analysis on the GenImage dataset. We report the ASR against SAFE, alongside visual quality metrics, by halving ($0.5\times$) and doubling ($2\times$) the default value of each hyperparameter.}
\label{tab:sensitivity}
\resizebox{\columnwidth}{!}{
\begin{tabular}{ll c cccc}
\toprule
\textbf{Parameter} & \textbf{Value} & \textbf{ASR} (\%)  & \textbf{PSNR}  & \textbf{SSIM}  & \textbf{$L_2$}($\times 10^{-2}$) & \textbf{FID}  \\
\midrule

\multirow{3}{*}{$\sigma_p$ (Poisson)} 
& 0.045 ($0.5\times$) & 88.49 & 31.66 & 0.85 & 2.62 & 16.55 \\
& 0.090 (Default) & 81.54 & 35.56 & 0.91 & 1.68 & 14.21 \\
& 0.180 ($2\times$) & 85.93 & 34.29 & 0.90 & 1.94 & 8.83 \\
\midrule

\multirow{3}{*}{$\sigma_g$ (Gaussian)} 
& 0.0375 ($0.5\times$) & 71.59 & 31.92 & 0.85 & 2.55 & 14.08 \\
& 0.0750 (Default) & 81.54 & 35.56 & 0.91 & 1.68 & 14.21 \\
& 0.1500 ($2\times$) & 84.69 & 32.34 & 0.86 & 2.44 & 13.16 \\
\midrule

\multirow{3}{*}{$\alpha$ (Blend Factor)} 
& 0.100 ($0.5\times$) & 59.18 & 33.33 & 0.89 & 2.25 & 12.14 \\
& 0.200 (Default) & 81.54 & 35.56 & 0.91 & 1.68 & 14.21 \\
& 0.400 ($2\times$) & 71.97 & 38.76 & 0.96 & 1.16 & 5.31 \\
\midrule

\multirow{3}{*}{$\epsilon$ (Bound)} 
& 0.003 ($0.5\times$) & 66.12 & 43.77 & 0.99 & 0.65 & 2.15 \\
& 0.006 (Default) & 81.54 & 35.56 & 0.91 & 1.68 & 14.21 \\
& 0.012 ($2\times$) & 98.91 & 30.55 & 0.82 & 3.04 & 12.45 \\
\midrule

\multirow{3}{*}{$\gamma$ (Mask Power)} 
& 1.250 ($0.5\times$) & 96.80 & 33.16 & 0.88 & 2.20 & 11.29 \\
& 2.500 (Default) & 81.54 & 35.56 & 0.91 & 1.68 & 14.21 \\
& 5.000 ($2\times$) & 62.88 & 38.07 & 0.95 & 1.29 & 7.03 \\
\bottomrule
\end{tabular}
}
\end{table}

\subsection{Ablation Study and Sensitivity Analysis}
\label{sec:ablation}

To rigorously evaluate the contribution of each module and the robustness of our framework to hyperparameter variations, we conduct comprehensive ablation studies on the GenImage dataset against the SAFE detector. 

\subsubsection{Component Analysis.} Table~\ref{tab:ablation} illustrates the impact of individual components in our pipeline. Starting from a naive noise addition baseline, we observe that purely injecting statistical noise without the ISP module yields a suboptimal ASR of 62.73\%. Integrating the physically grounded ISP module significantly elevates the ASR to 81.54\%. This validates our core motivation: simulating the nonlinear transformations of a real camera pipeline ensures the injected perturbations are statistically indistinguishable from authentic sensor artifacts, thereby effectively evading forensic detectors.

Furthermore, spatial and magnitude constraints prove essential for maintaining visual fidelity. Disabling the clipping boundary ($\epsilon$) leads to an abnormally high ASR of 99.26\%, but this comes at the catastrophic cost of severe visual distortion, evidenced by the FID skyrocketing to 121.43 and PSNR dropping to 13.93. Similarly, removing the adaptive masking strategy causes a noticeable drop in both ASR (74.04\%) and image quality. The mask modulates the high-frequency components of the perturbation, suppressing responses in smooth regions while preserving structured variations that are more consistent with natural image statistics. Ultimately, the full model achieves an optimal balance, ensuring high adversarial effectiveness without compromising imperceptibility.

\subsubsection{Sensitivity and Trade-off Analysis.} Table~\ref{tab:sensitivity} presents a sensitivity analysis of the major hyperparameters, halving ($0.5\times$) and doubling ($2\times$) their default values to analyze the system's behavior.

The noise scaling factors $\sigma_p$ and $\sigma_g$ exhibit comparatively stable behavior. Varying $\sigma_p$ and $\sigma_g$ over a wide range leads to only moderate changes in both ASR and image quality, without abrupt degradation. This is because the overall perturbation magnitude is explicitly constrained by the bound $\epsilon$, which prevents excessive noise amplification even when the noise levels increase.

In contrast, the blending factor $\alpha$ plays a more active role in controlling the attack strength. When $\alpha$ is too small, the injected perturbation is insufficient to effectively alter the detector response, leading to lower ASR. Increasing $\alpha$ improves attack effectiveness up to a certain point, but overly large values cause the perturbation to deviate from the underlying ISP-consistent structure, which weakens its ability to remain statistically aligned with natural image distributions and thus reduces ASR.

The mask power $\gamma$ primarily controls the artifact-removal step. Smaller $\gamma$ values allow broader perturbations, leading to higher ASR but slightly lower visual quality. Larger $\gamma$ focuses the perturbation, reducing ASR while improving perceptual metrics. Thus, $\gamma$ balances attack effectiveness with artifact suppression.

Overall, the sensitivity analysis reveals that $\epsilon$ and $\gamma$ dominate the trade-off between attack effectiveness and perceptual quality, while $\sigma_p$ and $\sigma_g$ are regulated by the perturbation bound and therefore have limited impact. The blending factor $\alpha$ introduces an additional balance between perturbation strength and structural consistency. Together, these observations explain how different components interact to produce stable and effective adversarial behavior.

\begin{table}[h]
\centering
\caption{Computational efficiency. We reported execution time in seconds for generating $1,000$ adversarial images.}
\label{tab:efficiency}
\resizebox{\columnwidth}{!}{
\begin{tabular}{l|ccccc}
\toprule
Method & StealthDiffusion & DiffAttack & Diff-PGD & PGD & ISPCloak (Ours) \\
\midrule
Time & 4,842 & 3,907 & 2,511 & 52 & 32 \\
\bottomrule
\end{tabular}
}
\end{table}

\subsection{Computational Efficiency}

We evaluate computational overhead by measuring the execution time on an NVIDIA RTX 4090D to generate $1,000$ adversarial examples on the GenImage ADM subset. As shown in Table~\ref{tab:efficiency}, diffusion-based attacks incur prohibitive costs due to iterative sampling. The gradient-based PGD is faster than existing methods, but it still requires multiple backpropagation steps. Conversely, ISPCloak leverages an optimization-free, single forward-pass pipeline, processing the entire batch in merely $32$ seconds. This ultra-fast generation speed enables highly practical, real-time deployment.

\section{Conclusion}
\label{sec:conclusion}

In this paper, we present ISPCloak, an adversarial attack framework that exposes vulnerabilities in modern AI-generated image detectors. Unlike traditional optimization-based attacks, ISPCloak leverages a reversible ISP pipeline to embed authentic sensor imprints into synthetic images. By injecting statistically consistent shot and read noise with adaptive spatial constraints, our method effectively conceals digital generative artifacts while preserving visual fidelity. Extensive evaluations on GenImage, WildFake, and FaceForensics++ demonstrate that ISPCloak consistently achieves high ASR across heterogeneous detectors under strict black-box conditions. Moreover, the manipulated images become statistically indistinguishable from real camera captures, highlighting a critical blind spot in current forensic defenses. These findings underscore the need for robust, physics-aware detection mechanisms capable of resisting ISP-consistent perturbations.

\bibliographystyle{ACM-Reference-Format}
\bibliography{main}

\end{document}